\documentclass[runningheads]{llncs}
\usepackage{graphicx}
\usepackage{amsmath,amssymb} 
\usepackage{color}

\usepackage{color}
\usepackage{comment}
\usepackage{epsfig}
\usepackage{graphicx}
\usepackage{booktabs}
\usepackage{colortbl}
\usepackage{subcaption}

\usepackage[numbers,sort,compress]{natbib}

\begin{document}
\pagestyle{headings}
\mainmatter
\def\ACCV20SubNumber{752}  

\title{Uncertainty Estimation and Sample Selection \\ for Crowd Counting} 
\titlerunning{Uncertainty Estimation \& Sample Selection for Crowd Counting}
%
\author{Viresh Ranjan\inst{1}
\and
Boyu Wang\inst{1}
\and
Mubarak Shah\inst{2}
\and 
Minh Hoai\inst{1}
}
\authorrunning{Ranjan et al.}
%
\institute{Department of Computer Science, Stony Brook University, Stony Brook, NY 11790 \and 
University of Central Florida, Orlando, FL 32816
}

\maketitle


\begin{abstract}
We present a method for image-based crowd counting, one that can predict a crowd density map together with the uncertainty values pertaining to the predicted density map. To obtain prediction uncertainty, we model the crowd density values using Gaussian distributions and develop a convolutional neural network architecture to predict these distributions. A key advantage of our method over existing crowd counting methods is its ability to quantify the uncertainty of its predictions. We illustrate the benefits of knowing the prediction uncertainty by developing a method to reduce the human annotation effort needed to adapt counting networks to a new domain. We present sample selection strategies which make use of the density and uncertainty of predictions from the networks trained on one domain to select the informative images from a target domain of interest to acquire human annotation. We show that our sample selection strategy drastically reduces the amount of labeled data from the target domain needed to adapt a counting network trained on a source domain to the target domain. Empirically, the networks trained on the UCF-QNRF dataset can be adapted to surpass the performance of the previous state-of-the-art results on NWPU dataset and Shanghaitech dataset using only 17$\%$ of the labeled training samples from the target domain. 

Code: \url{https://github.com/cvlab-stonybrook/UncertaintyCrowdCounting}
\end{abstract}

\def\mA{\mathcal{A}}
\def\mB{\mathcal{B}}
\def\mC{\mathcal{C}}
\def\mD{\mathcal{D}}
\def\mE{\mathcal{E}}
\def\mF{\mathcal{F}}
\def\mG{\mathcal{G}}
\def\mH{\mathcal{H}}
\def\mI{\mathcal{I}}
\def\mJ{\mathcal{J}}
\def\mK{\mathcal{K}}
\def\mL{\mathcal{L}}
\def\mM{\mathcal{M}}
\def\mN{\mathcal{N}}
\def\mO{\mathcal{O}}
\def\mP{\mathcal{P}}
\def\mQ{\mathcal{Q}}
\def\mR{\mathcal{R}}
\def\mS{\mathcal{S}}
\def\mT{\mathcal{T}}
\def\mU{\mathcal{U}}
\def\mV{\mathcal{V}}
\def\mW{\mathcal{W}}
\def\mX{\mathcal{X}}
\def\mY{\mathcal{Y}}
\def\mZ{\mathcal{Z}}

\def\1n{\mathbf{1}_n}
\def\0{\mathbf{0}}
\def\1{\mathbf{1}}

\def\A{{\bf A}}
\def\B{{\bf B}}
\def\C{{\bf C}}
\def\D{{\bf D}}
\def\E{{\bf E}}
\def\F{{\bf F}}
\def\G{{\bf G}}
\def\H{{\bf H}}
\def\I{{\bf I}}
\def\J{{\bf J}}
\def\K{{\bf K}}
\def\L{{\bf L}}
\def\M{{\bf M}}
\def\N{{\bf N}}
\def\O{{\bf O}}
\def\P{{\bf P}}
\def\Q{{\bf Q}}
\def\R{{\bf R}}
\def\S{{\bf S}}
\def\T{{\bf T}}
\def\U{{\bf U}}
\def\V{{\bf V}}
\def\W{{\bf W}}
\def\X{{\bf X}}
\def\Y{{\bf Y}}
\def\Z{{\bf Z}}

\def\a{{\bf a}}
\def\b{{\bf b}}
\def\c{{\bf c}}
\def\d{{\bf d}}
\def\e{{\bf e}}
\def\f{{\bf f}}
\def\g{{\bf g}}
\def\h{{\bf h}}
\def\i{{\bf i}}
\def\j{{\bf j}}
\def\k{{\bf k}}
\def\l{{\bf l}}
\def\m{{\bf m}}
\def\n{{\bf n}}
\def\o{{\bf o}}
\def\p{{\bf p}}
\def\q{{\bf q}}
\def\r{{\bf r}}
\def\s{{\bf s}}
\def\t{{\bf t}}
\def\u{{\bf u}}
\def\v{{\bf v}}
\def\w{{\bf w}}
\def\x{{\bf x}}
\def\y{{\bf y}}
\def\z{{\bf z}}

\def\balpha{\mbox{\boldmath{$\alpha$}}}
\def\bbeta{\mbox{\boldmath{$\beta$}}}
\def\bdelta{\mbox{\boldmath{$\delta$}}}
\def\bgamma{\mbox{\boldmath{$\gamma$}}}
\def\blambda{\mbox{\boldmath{$\lambda$}}}
\def\bsigma{\mbox{\boldmath{$\sigma$}}}
\def\btheta{\mbox{\boldmath{$\theta$}}}
\def\bomega{\mbox{\boldmath{$\omega$}}}
\def\bxi{\mbox{\boldmath{$\xi$}}}
\def\bnu{\mbox{\boldmath{$\nu$}}}                                  
\def\bphi{\mbox{\boldmath{$\phi$}}}
\def\bmu{\mbox{\boldmath{$\mu$}}}

\def\bDelta{\mbox{\boldmath{$\Delta$}}}
\def\bOmega{\mbox{\boldmath{$\Omega$}}}
\def\bPhi{\mbox{\boldmath{$\Phi$}}}
\def\bLambda{\mbox{\boldmath{$\Lambda$}}}
\def\bSigma{\mbox{\boldmath{$\Sigma$}}}
\def\bGamma{\mbox{\boldmath{$\Gamma$}}}

\newcommand{\myminimum}[1]{\mathop{\textrm{minimum}}_{#1}}
\newcommand{\mymaximum}[1]{\mathop{\textrm{maximum}}_{#1}}    
\newcommand{\mymin}[1]{\mathop{\textrm{minimize}}_{#1}}
\newcommand{\mymax}[1]{\mathop{\textrm{maximize}}_{#1}}
\newcommand{\mymins}[1]{\mathop{\textrm{min.}}_{#1}}
\newcommand{\mymaxs}[1]{\mathop{\textrm{max.}}_{#1}}  
\newcommand{\myargmin}[1]{\mathop{\textrm{argmin}}_{#1}} 
\newcommand{\myargmax}[1]{\mathop{\textrm{argmax}}_{#1}} 
\newcommand{\myst}{\textrm{s.t. }}

\newcommand{\denselist}{\itemsep -1pt}
\newcommand{\sparselist}{\itemsep 1pt}

\definecolor{pink}{rgb}{0.9,0.5,0.5}
\definecolor{purple}{rgb}{0.5, 0.4, 0.8}   
\definecolor{gray}{rgb}{0.3, 0.3, 0.3}
\definecolor{mygreen}{rgb}{0.2, 0.6, 0.2}

\newcommand{\cyan}[1]{\textcolor{cyan}{#1}}
\newcommand{\red}[1]{\textcolor{red}{#1}}  
\newcommand{\blue}[1]{\textcolor{blue}{#1}}
\newcommand{\magenta}[1]{\textcolor{magenta}{#1}}
\newcommand{\pink}[1]{\textcolor{pink}{#1}}
\newcommand{\green}[1]{\textcolor{green}{#1}} 
\newcommand{\gray}[1]{\textcolor{gray}{#1}}    
\newcommand{\mygreen}[1]{\textcolor{mygreen}{#1}}    
\newcommand{\purple}[1]{\textcolor{purple}{#1}}       

\definecolor{greena}{rgb}{0.4, 0.5, 0.1}
\newcommand{\greena}[1]{\textcolor{greena}{#1}}

\definecolor{bluea}{rgb}{0, 0.4, 0.6}
\newcommand{\bluea}[1]{\textcolor{bluea}{#1}}
\definecolor{reda}{rgb}{0.6, 0.2, 0.1}
\newcommand{\reda}[1]{\textcolor{reda}{#1}}

\def\changemargin#1#2{\list{}{\rightmargin#2\leftmargin#1}\item[]}
\let\endchangemargin=\endlist
                                               
\newcommand{\cm}[1]{}

\newcommand{\mtodo}[1]{{\color{red}$\blacksquare$\textbf{[TODO: #1]}}}
\newcommand{\myheading}[1]{\vspace{1ex}\noindent \textbf{#1}}
\newcommand{\htimesw}[2]{\mbox{$#1$$\times$$#2$}}
\newcommand{\mh}[1]{\textcolor{blue}{[Minh: {#1}]}}
\newcommand{\ms}[1]{\textcolor{red}{[MS: {#1}]}}

\newif\ifshowsolution
\showsolutiontrue

\ifshowsolution  
\newcommand{\Comment}[1]{\paragraph{\bf $\bigstar $ COMMENT:} {\sf #1} \bigskip}
\newcommand{\Solution}[2]{\paragraph{\bf $\bigstar $ SOLUTION:} {\sf #2} }
\newcommand{\Mistake}[2]{\paragraph{\bf $\blacksquare$ COMMON MISTAKE #1:} {\sf #2} \bigskip}
\else
\newcommand{\Solution}[2]{\vspace{#1}}
\fi

\newcommand{\truefalse}{
\begin{enumerate}
	\item True
	\item False
\end{enumerate}
}

\newcommand{\yesno}{
\begin{enumerate}
	\item Yes
	\item No
\end{enumerate}
}
\newcommand{\Sref}[1]{Sec.~\ref{#1}}
\newcommand{\Eref}[1]{Eq.~(\ref{#1})}
\newcommand{\Fref}[1]{Fig.~\ref{#1}}
\newcommand{\Tref}[1]{Tab.~\ref{#1}}

\definecolor{customgray}{rgb}{0.9, 0.9, 0.9}
\newcolumntype{d}{>{\columncolor[gray]{.7}}c}
\newcolumntype{g}{>{\columncolor[gray]{.9}}c}
\newcolumntype{z}{>{\columncolor[gray]{.9}}l}
  
\section{Introduction}\label{sec:Introduction}

Crowd counting from unconstrained images is a challenging task due to the large variation in occlusion, crowd density, and camera perspective. Most  recent methods~\cite{zhang2016single,sindagi2017generating,sindagi2017cnn,sam2017switching,ranjan2018iterative,bayesianCounting,wang2019learning,ranjan2019crowd}  learn a Convolutional Neural Network (CNN) to map an input image to the corresponding crowd density map, from which the total count can be computed by summing all the predicted density values at all pixels. Although the performance of the crowd counting methods have improved significantly in the recent years, their performance on the challenging datasets such as~\cite{idrees2018composition,zhang2016single,idrees2013multi} is far below the human-level performance.  One factor affecting the performance of the existing crowd counting systems is the limited amount of annotated training data. The largest crowd counting dataset~\cite{wang2020nwpu} consist of 5,109 images. Annotating dense crowd images, which involves placing a dot over the head of each person in the crowd, is time consuming. This makes it harder to create large-scale crowd counting datasets.
\begin{figure}[!t]  
\centering
\begin{subfigure}[b]{0.31\textwidth}
\includegraphics[width=\textwidth]{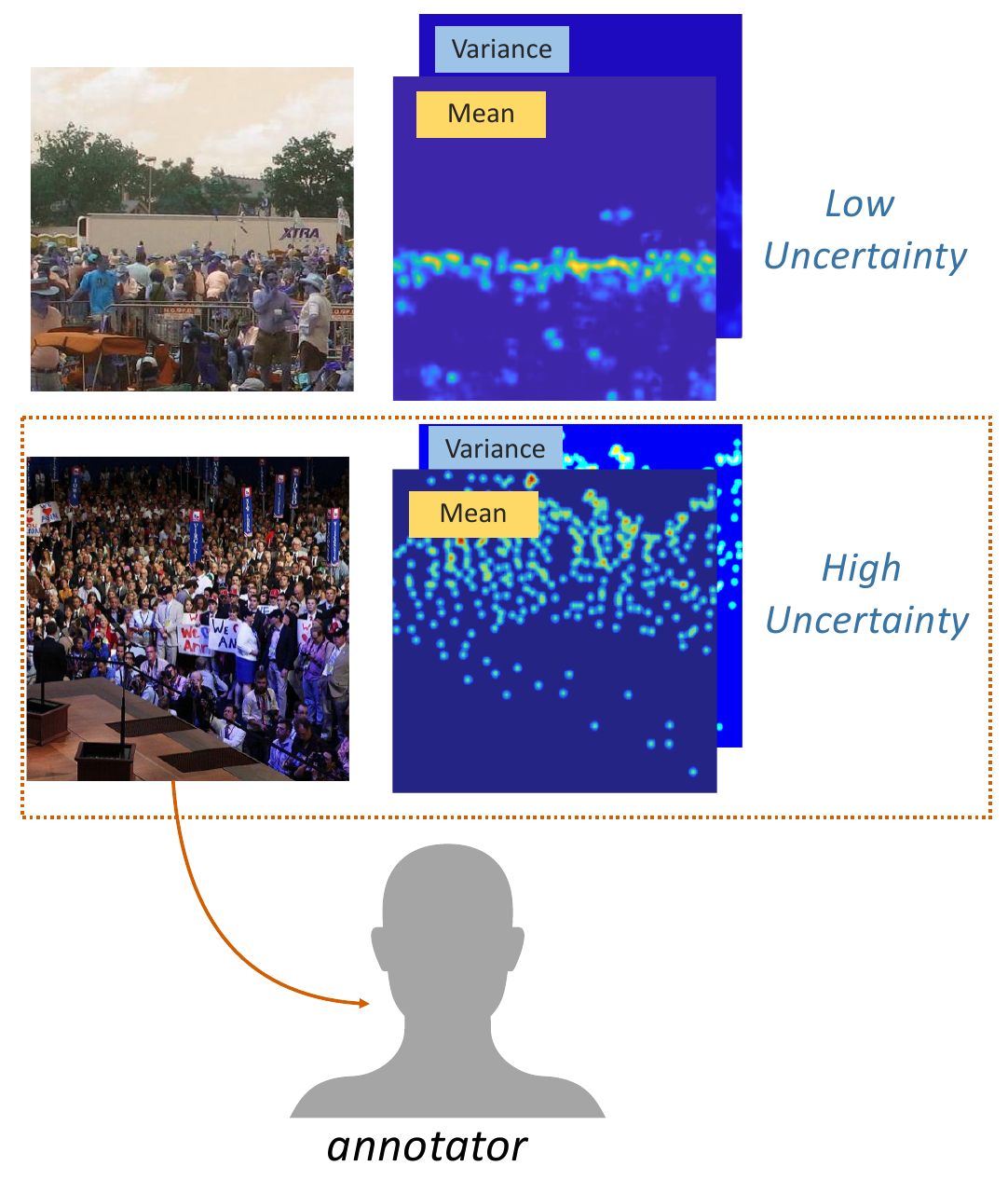}
\caption{\label{fig:select_policy_1}}
\end{subfigure} 
\begin{subfigure}[b]{0.45\textwidth}
\includegraphics[width=\textwidth]{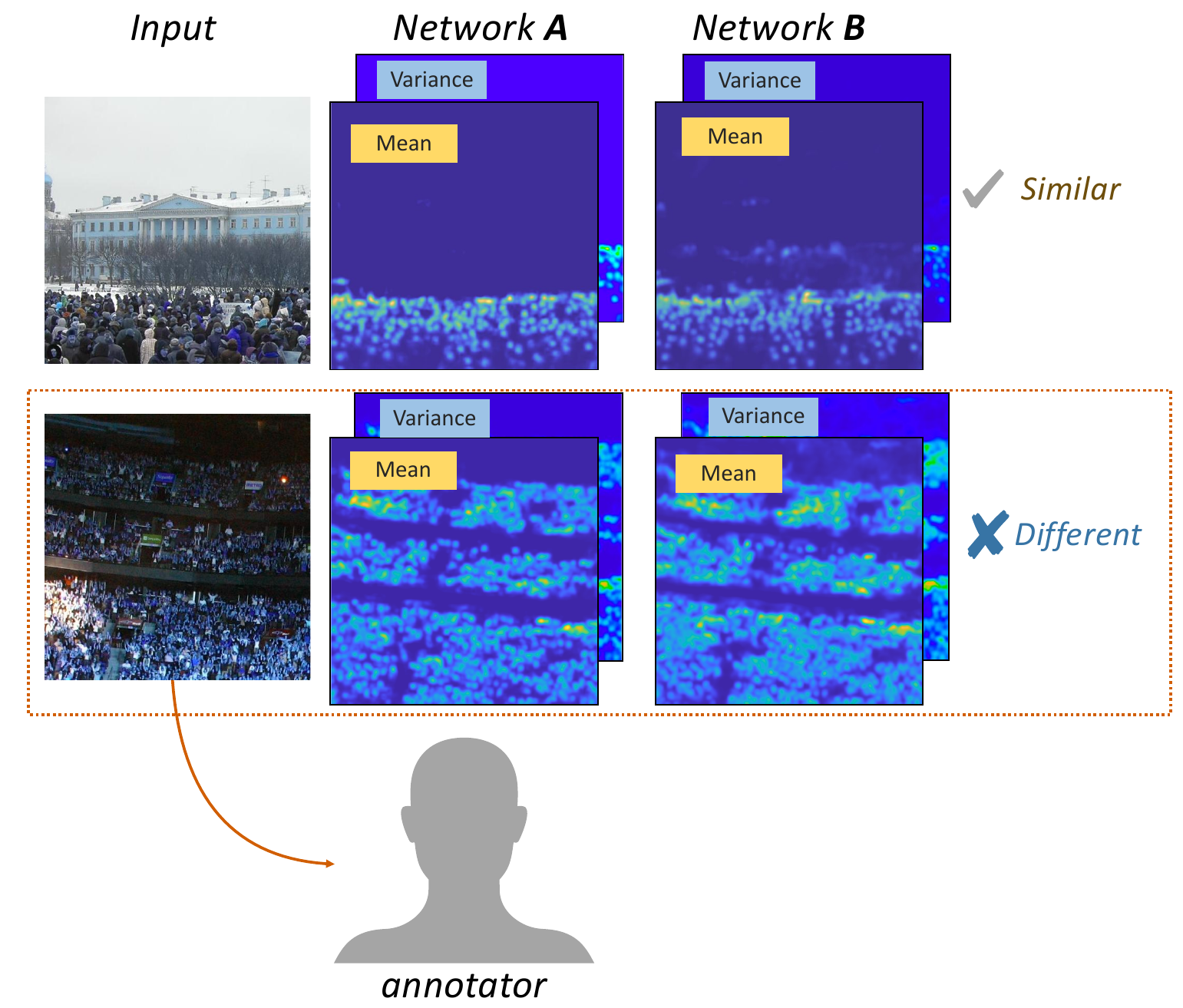}
\caption{\label{fig:select_policy_2}}
\end{subfigure} 
	\caption{
     \textbf{Different sample selection strategies based on uncertainty estimation}. (a) uncertainty based sample selection: Images with higher average uncertainty values are selected for annotation. (b) Ensemble disagreement based sample selection:  Given networks A and B trained on a source domain, and a set of unlabeled images from a target domain, we obtain the crowd density map and uncertainty values from both networks for all images in the target domain. Based on the prediction, we compute the disagreement between the two networks. Images with large disagreement are picked for human annotation.
     \label{fig:SampleSelection}}	
\end{figure}

In this paper, we present an approach to tackle the prohibitively large costs involved in annotating crowd images. Our approach draws inspiration from Active Learning, which is based on the hypothesis that a learning algorithm can perform well with less training data if it is allowed to select \textit{informative samples}~\cite{lewis1994sequential,settles2009active}. Given a pool of labeled crowd images from the source domain and a large pool of unlabeled crowd images, we are interested in identifying a subset of informative samples from the unlabeled pool, and instead of annotating the whole pool, we obtain the annotation for these selected samples. 
To find most informative samples from the unlabeled pool, we train networks on the labeled pool first. Next, we select those samples from the unlabeled pool for which the networks are uncertain about their predictions.  However, most existing crowd counting  methods do not provide any measure of uncertainty for their predictions. 
We develop a fully convolutional network architecture for estimating the crowd density and the corresponding uncertainty of prediction. For uncertainty estimation, we follow the approach of Nix and Weigand \cite{nix1994estimating} who used a Multi-Layer Perceptron (MLP) to estimate the uncertainty of prediction. This approach assumes that observed output values are drawn from a Gaussian distribution, and the MLP predicts the mean and variance of the Gaussian distribution. The network is trained by maximizing the log likelihood of the data. The variance serves as a measure of uncertainty of the prediction. 

Inspired by Nix and Weigand~\cite{nix1994estimating}, we develop a fully convolutional architecture with a shared trunk for feature extraction and two prediction heads for predicting the crowd density map and the corresponding variance. This network is trained on the source domain by maximizing the log likelihood of the data. We use the predictions from this network, and present two sampling strategies for selecting informative samples from the target domain for human annotation. We present the overview of our sampling strategy in \Fref{fig:SampleSelection}. Our sampling strategy can be used for selecting images from a large pool of unlabeled images, and it can also be used to pick informative crops from an image. Depending on the annotation budget, it might be useful to get partial annotations for an image by picking informative crops from an image and getting human annotations for the informative crops rather than annotating the entire image. We present experiments on image level sample selection\footnote{Experiments on crop sample selection are presented in the supplementary materials.} and empirically show that the networks trained on UCF-QNRF dataset can be adapted to surpass the performance of the previous state-of-the-art results on NWPU dataset using less than 17$\%$ of the labeled training samples. We also show that the UCF-QNRF pretrained networks can be adapted to perform well on the Shanghaitech dataset as well, with only a third of annotated examples from Shanghaitech dataset. Our results clearly show the usefulness of using our sampling strategy in saving human annotation cost, and it can help reduce human annotation cost involved with annotating large scale crowd datasets. Our sampling strategy isn't specific to crowd counting, and it can be applied to any other pixel level prediction task such as optical flow estimation, semantic segmentation as well. We decide to focus on Crowd Counting in this paper since human annotation is particularly expensive for Crowd Counting.

The main contributions of our work are: \textbf{(1)} We propose a novel network architecture for crowd density prediction and corresponding uncertainty estimation that uses both local features and self-attention based non-local features for prediction. \textbf{(2)} We show that modeling prediction uncertainty leads to a more robust loss function, which outperforms the commonly used mean squared loss, obtaining state of the art results on multiple crowd counting datasets. 
\textbf{(3)} We present a novel uncertainty guided sample selection strategy that enables using networks trained on one domain to select informative samples from another domain for human annotation. To the best of our knowledge, ours is the first work focusing on using predictive uncertainty for sample selection pertaining to any pixel level prediction task in Computer Vision. We show empirically that using the proposed sampling strategy, it is possible to adapt a network trained on a source domain to perform well on the target domain using significantly less annotated data from the target domain. 

\section{Related Work}
\label{sec:RelatedWork}

Crowd counting is an active area of research with two general approaches:  detection  approach and density estimation approach. Despite lots of related works,  none of them use non-local features to reduce the ambiguity of the estimation, nor use the uncertainty estimates for sample selection.

\myheading{Detection and Regression Based Approaches.}
Crowd counting has been studied for a long time. Initial crowd counting approaches~\cite{lin2001estimation,li2008estimating} were based on detection, which used a classifier such as SVMs trained on top of hand crafted feature representation. These approaches performed well on simpler crowd counting datasets, but their performance was severely affected by occlusion, which is quite common in dense crowd datasets. Some of the later  approaches~\cite{chan2009bayesian,chen2012feature} tried to tackle the occlusion problem by avoiding the detection problem, and directly learning a regression function to predict the count. 

\myheading{Density Estimation Based Approaches.}
The precursor to the current density estimation based approaches was the work of
\citet{lempitsky2010learning}, who presented one of the earliest works on density estimation based crowd counting. 
In the recent deep learning era, density estimation has become the de facto strategy for most of the recent crowd counting approaches
~\cite{zhang2016single,sam2017switching,li2018csrnet,onoro2016towards,ranjan2018iterative,idrees2018composition,cao2018scale,bayesianCounting,wang2019learning,liu2019context,shi2019revisiting,liu2019point,liu2019adcrowdnet,Xu_2019_ICCV,Cheng_2019_ICCV,Liu_2019_ICCV,Yan_2019_ICCV,Zhao_2019_CVPR,Zhang_2019_CVPR,Jiang_2019_CVPR,Wan_2019_CVPR,Lian_2019_CVPR,Liu_2019_CVPR, lu2018class}. 

Starting with \citet{zhang2016single}, many approaches~\cite{Sam-etal-CVPR17,ranjan2018iterative,sindagi2017generating} used multiple parallel feature convolutional columns to extract features at different resolution. \citet{zhang2016single} used a multi-column architecture comprising of three columns to address the large variation in crowd size and density.  The different columns had kernels of varying sizes.  The column with larger kernels could extract features for less dense crowd, while the column with finer kernels is for denser crowd. \citet{sam2017switching} proposed to  decouple different columns, and train them separately. Each column was specialized towards a certain density type. This made the task of each regressor easier, since it had to handle similar density images. They also trained a switch classifier which routed an image patch to the appropriate regressor. However, unlike \cite{zhang2016single}, the training procedure comprised of multiple stages. \citet{sindagi2017generating} presented an approach which utilized local and global context information for crowd counting. They trained a classifier for classifying an image into multiple density categories, and the classifier score was used to create context feature maps. 
\citet{ranjan2018iterative} used a two stage coarse to fine approach to predict crowd density map. In the first stage, a low resolution density map was predicted, which was later utilized as a feature map while predicting the final high resolution density map. 

\myheading{Uncertainty Estimation}
For Computer Vision tasks, we typically  consider two types of uncertainty: \textit{aleatoric} and \textit{epistemic}~\cite{kendall2017uncertainties}. Aleatoric uncertainty captures the uncertainty inherent in the data, and can be modeled by predicting the parameters of a Gaussian distribution and maximizing the log likelihood~\cite{nix1994estimating,kendall2017uncertainties} of the observed data. Epsitemic uncertainty, also called model uncertainty, is related to the uncertainty in the model parameters, and can be explained away given a large enough dataset. Epistemic uncertainity can be captured by Bayesian Neural Networks~\cite{neal2012bayesian}. Although performing inference with earlier Bayesian Neural Networks was inefficient, recent techniques like Monte Carlo Dropout~\cite{gal2015bayesian} can be used to capture epistemic uncertainty even with large neural networks. Some of the earlier works have focused on uncertainty prediction for tasks such as optical flow estimation~\cite{ilg2018uncertainty} and crowd counting~\cite{oh2019crowd}. However, none of these earlier works have focused on using uncertainty estimates for sample selection.



\section{Uncertainty Estimation for Crowd Counting}\label{sec:Uncertainty Estimation for Crowd Counting}
We take motivation from earlier work \citet{kendall2017uncertainties} which shows the usefulness of both Aleatoric and Epistemic Uncertainty estimates for various Computer Vision tasks, and present architectures which can be used for obtaining the two types of uncertainties. In \Sref{sec:Non-local Crowd Counting}, we present our proposed network architecture for estimating the aleatoric uncertainty\footnote{See supplementary materials for the architecture for estimating epistemic uncertainty}, followed by training objective in \Sref{sec:TrainingObjective}.

\subsection{Crowd Transformer Network\label{sec:Non-local Crowd Counting}}

In this section, we describe our Crowd Transformer Network (CTN) architecture, which predicts the crowd density map along with the corresponding uncertainty values.
CTN models the predictive uncertainty, i.e., the uncertainty inherent in the input image which might arise from sensor noise, or from the ambiguity in the image itself. 

\begin{figure*}[!t]
\centering
\includegraphics[width=\linewidth]{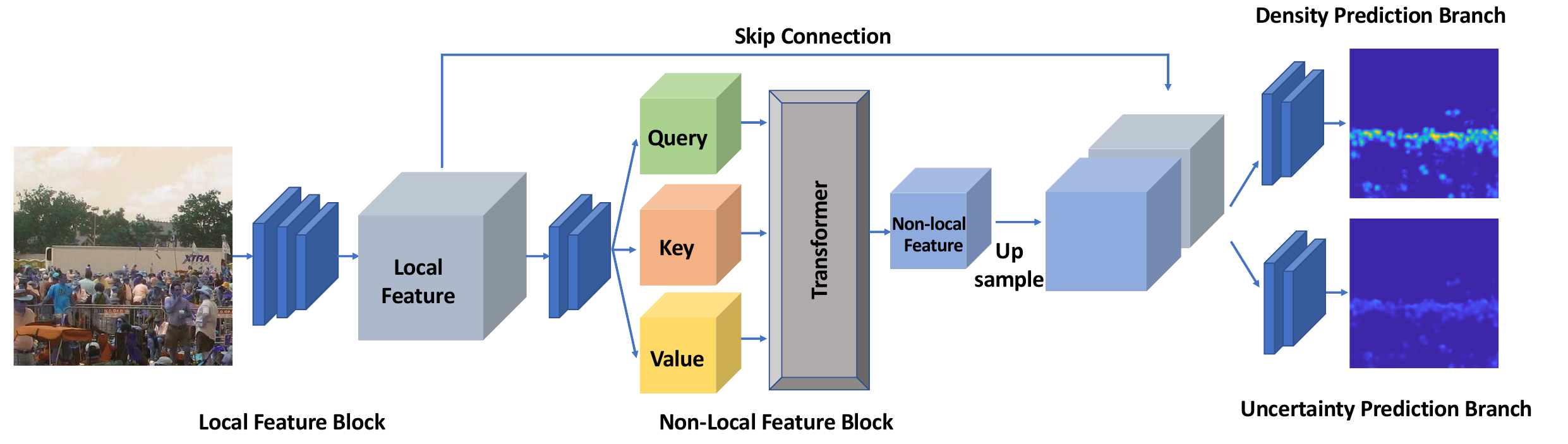}
	\caption{
     \textbf{CTN architecture} predicts both crowd density map and corresponding uncertainty values. It combines local and non-local features.
     The local features are computed by the convolution layers in the local feature block. The resulting feature map is passed to the non-local feature block. Both density prediction branch and uncertainty prediction branch utilize local and non-local features.
     \label{fig:CTN}}	
\end{figure*}

The block diagram of CTN is presented in \Fref{fig:CTN}. CTN uses both local and non-local features for estimating the crowd density map. 
 Let $X$ be a crowd image of size $H{\times}W$ and $Y$ the corresponding ground truth density map of the same size. We assume that each value in $Y$ is generated by a Gaussian distribution, and CTN predicts the mean and the variance of the Gaussian distribution. The proposed CTN 
takes input $X$ and 
 predicts mean and variance maps as:
 \begin{equation}
X \rightarrow \mu(X,\theta),\sigma^2(X,\theta)
\end{equation}
where $\mu(X,\theta)$ is the crowd density map and $\sigma^2(X,\theta)$ the uncertainty map. We use uncertainty and variance interchangeably in the rest of the paper. Both $\mu(X,\theta)$ and $\sigma^2(X,\theta)$ have the same size as the input image $X$ as well as the crowd density map $Y$. 
The key components of CTN architecture are: 1) \textit{a local feature block}, 2) \textit{a non-local feature block}, 3) \textit{a density prediction branch}, and 4) \textit{a uncertainty prediction branch}. These components are described below.

\myheading{Local Feature Block.}\label{sec:LocalFeatureBlock}
Given an input image $X$ of size $H{\times}W$, we pass it through a local feature block to obtain the convolutional feature maps. The local feature block consists of five convolution layers with kernels of size $3{\times}3$, and the number of filters in the convolution  layers are 64, 64, 128, 128, and 256. We use the VGG16 network~\cite{simonyan2014very} pretrained on ImageNet to initialize the convolution layers in the local feature block. The local feature block has two max pooling layers, after the second and fourth convolution layers. The resulting feature map is a tensor of size $\frac{H}{4}{\times}\frac{W}{4}{\times}256$. The feature map is passed to the non-local block as well as the density prediction branch and uncertainty branch. 

\myheading{Non-local Feature Block.}\label{sec:NonLocalFeatureBlock}
For computing the non-local features, we use the Transformer architecture~\cite{vaswani2017attention}
which was proposed as an alternative to recurrent neural network~\cite{Rumelhart-et-al-PDP86} for solving various sequence analysis  tasks. It uses an attention mechanism to encode non-local features. The architecture consists of an encoder and a decoder, where the encoder maps the input sequence into an intermediate representation, which in turn is mapped by the decoder into the output sequence. The transformer uses three types of attention layers: \textit{encoder self-attention, encoder-decoder attention}, and \textit{decoder self-attention}. For the proposed crowd counting approach in this paper, only the first one is relevant which we describe briefly next. Henceforth, we will use self-attention  to refer to the self-attention of the encoder. 

\myheading{Encoder Self-Attention}. Given a query sequence along with a key-and-value sequence, the self-attention layer outputs a sequence where the $i$-th element in the output sequence is obtained as a weighted average of the value sequence, and the weights are decided based on the similarity between the $i$-th query element and the key sequence. Let $X \in R^{n\times d}$ be a matrix representation for a sequence consisting of $n$ vectors of $d$ dimensions.
The self-attention layer first transforms $X$ into query $X_Q$, key $X_K$, and value $X_V$ matrices by multiplying $X$ with matrices $W_Q$, $W_K$, and $W_V$, respectively:
\begin{align}
 X_Q = X W_Q, X_K = X W_K, X_V = X W_V.
\end{align}
The output sequence $Z$ is computed efficiently with matrix multiplications:
\begin{equation}\label{eqn:eqn2}
 Z = softmax(X_Q X_K^T)X_V.   
\end{equation}
The encoder consists of multiple self-attention layers, arranged in a sequential order so that the output of one layer is fed as input to the next layer. 

\myheading{Architecture Details.} The non-local feature block takes as input the feature map from the local feature block, and passes it through three convolution layers of kernel size $3{\times}3$ and a max pooling layer, which results in a feature map of size $\frac{H}{8}{\times}\frac{W}{8}{\times}512$. We reduce the depth of the feature map by passing it through a $1{\times}1$ convolution layer, which yields a feature map of size $\frac{H}{8}{\times}\frac{W}{8}{\times}240$. The resulting feature map is flattened into a matrix of size $M{\times}240$, where $M = \frac{H}{8}{\times}\frac{W}{8}$. Each row in this matrix corresponds to some location in the convolution feature map. The flattened matrix is passed through three self-attention layers. The output from final transformer layer is reshaped back into a tensor of size $\frac{H}{8}{\times}\frac{W}{8}{\times}240$. 

\myheading{Density Prediction Branch.}\label{sec:DensityPredictionHead}
Both local and non-local features are important for estimating an accurate crowd density map. Hence, the Density Prediction Branch uses a skip connection to obtain the convolutional features from the local feature block, and combines it with the features from the non-local feature block. The non-local features are upsampled to the same spatial size as local features, which results in a tensor of size $\frac{H}{4}{\times}\frac{W}{4}{\times}240$. The local and non-local features are concatenated and passed through four convolution layers (with 196, 128, 48, and 1 filters), where the last layer is a $1{\times}1$ convolution layer. We add a ReLU non-linearity after the $1{\times}1$ convolution layer to prevent the network from predicting negative density values. We use two bilinear interpolation layers, after the second and third convolution layers in the prediction head. Each interpolation layer upsamples its input by the factor of two.  The input to the final $1{\times1}$ convolution layer is a feature map of size $H{\times}W{\times}48$, which is transformed into a 2D map by the last convolution layer.

\myheading{Predictive Uncertainty Estimation Branch.}\label{sec:UncertaintyPredictionHead}
The \textsl{Predictive Uncertainty Estimation Branch} outputs the variance map $\sigma^2(X,\theta)$. Similar to the density prediction branch, the uncertainty branch also uses both the local and non-local features for prediction. The uncertainty prediction branch has the same architecture as the density prediction branch, with one major difference being that we use point-wise softplus nonlinearity instead of ReLU nonlinearity after the last~$1{\times}1$ convolution layer. Softplus nonlinearity can be expressed as:
$softplus(x) = \frac{1}{\beta} \log(1 + \exp(\beta  x))$. For brevity, we will refer to this type of uncertainty estimation as Predictive Uncertainty.

\subsection{Training Objective}\label{sec:TrainingObjective}
The network is trained by minimizing the negative conditional log likelihood of the observed ground truth density values~$Y$, conditioned on the input image $X$:
 \begin{equation}
\mL(Y | X, \theta) = -\sum_{i=1}^{HW} \log( \mathbb{P}(Y_{i}|\mu_{i}(X,\theta),\sigma_{i}^2(X,\theta))),
\end{equation}
where $\mathbb{P}(y|\mu, \sigma^2) = \frac{1}{\sqrt{2\pi\sigma^2}}\exp(-\frac{(y - \mu)^2}{2\sigma^2})$, a univariate Gaussian distribution. The negative conditional log likelihood is proportional to: 
 \begin{equation} 
\mL(Y|X, \theta) \propto \sum_{i=1}^{HW} \left(\log{\sigma_{i}(X, \theta)} + \frac{(Y_{i} - \mu_{i}(X, \theta))^2}{2\sigma^2_{i}(X, \theta)}\right). \nonumber  
\end{equation}
The above objective can be seen as a weighted sum of the squared differences, where the weights depend on the estimated uncertainty of the input $X$. This objective can be seen as a robust regression objective, where higher importance is given to pixels with lower ambiguity~\cite{nix1994estimating}.


\section{Uncertainty Guided Sample Selection}\label{sec:Uncertainty Guided Sample Selection}
Given a labeled dataset $\{(X_A,Y_A)\}$ from domain $A$, and an unlabeled dataset $\{X_B\}$ from domain $B$, we are interested in finding a small subset of informative samples from domain $B$. Each instance $X_B$ of domain $B$  can be sent to an oracle (human) to obtain the label $Y_{B}$. Our motivation behind selecting a small subset from domain $B$ is to reduce the human annotation cost without sacrificing the performance on domain $B$. Next, we propose different strategies for selecting informative samples. In \Sref{sec:aleatoricSampling}, we propose to use the aleatoric uncertainty predicted by CTN to select informative samples. In \Sref{sec:EnsembleSampling}, we draw inspiration from Query-by-committee~\cite{settles2009active} sampling strategies in Active Learning, and present a sampling strategy that uses the disagreement between the members of an ensemble of CTN networks for selecting informative samples. We present two strategies for computing the disagreement, the first one uses both the density and the uncertainty predictions while the other uses just the density prediction.

\subsection{Aleatoric Uncertainty Based Sample Selection}\label{sec:aleatoricSampling}
When picking samples from the target domain, we want to select those samples for which the network makes erroneous prediction. Previous works and our own experiments show that aleatoric uncertainty is correlated to the prediction error (see supplementary materials and \Fref{fig:qualiContext}). Hence, we propose to use the aleatoric uncertainty for selecting informative samples. We use the CTN network trained on the source domain to compute the aleatoric uncertainty (averaged across the image) for all the images in the target domain, and select those samples from the target domain for labeling that have a high average aleatoric uncertainty.

\subsection{Ensemble Disagreement based Sample Selection}\label{sec:EnsembleSampling}
Inspired by the Query-By-Committee~\cite{seung1992query} sampling algorithm in Active Learning, we present another sampling strategy which uses the predictions from our CTN network trained on a source domain to select informative samples from a target domain.
The Query-By-Committee algorithm keeps a committee of students, and picks the sample with maximal disagreement between the committee members to acquire annotation. In this work, the committee is a set of two CTN networks as described in the previous section. These networks are trained on different subsets of labeled data from domain A, and the disagreement between the two networks are used as a measure of informativeness. Let the two networks be represented by their parameters $\theta_1$ and $\theta_2$ and the outputs of the networks are the mean and variance maps: 
 \begin{equation}\label{eqn:eqn5}
\left[\mu(X,\theta_1),\sigma^2(X,\theta_1)\right] \ \textrm{and }   \left[\mu(X,\theta_2),\sigma^2(X,\theta_2) \right].
\end{equation}
The values $\mu_{i}(X, \theta_1)$ and $\sigma_{i}^2(X, \theta_1)$ are the mean and variance of a Gaussian distribution for the density value at pixel $i$. Similarly, the values $\mu_{i}(X, \theta_2)$ and $\sigma_{i}^2(X, \theta_2)$ correspond to another Gaussian distribution. We use the KL divergence between these two distributions as a measure of disagreement between the two density estimation networks. We denote the KL divergence at location $i$ of image $X$ as $KL(X_i)$, which can be computed in close form as:
 \begin{align}\label{eqn:eqn6}
KL(X_i) = &  \frac{\sigma_{i}^2(X, \theta_1) + (\mu_{i}(X, \theta_1)-\mu_{i}(X, \theta_2))^2}{2\sigma_{i}^2(X, \theta_2)} 
+ \log\left(\frac{\sigma_{i}(X, \theta_2)}{\sigma_{i}(X, \theta_1)}\right) - \frac{1}{2}\cdot
\end{align}  
The overall informativeness of an image is obtained by computing the average KL divergence over all pixels. We sort all the images in domain B according to their informativeness, and select the most informative samples for annotation. Note that this approach can be easily extended for more than two networks.

We present another strategy called \textit{Density-difference based Ensemble disagreement} to compute the disagreement between the members of an ensemble. This disagreement is computed by averaging the pixel wise squared difference between the the density maps predicted by the members of the ensemble as
 \begin{align}
Diff(X_i) =    (\mu_{i}(X, \theta_1)-\mu_{i}(X, \theta_2))^2.
\end{align} 
The informativeness score is obtained by averaging $Diff(X_i)$ over the entire prediction map. The score can be  generalized to work with an ensemble of multiple networks.

\section{Experiments}
\label{sec:Results}
We validate the proposed approach by conducting experiments on four publicly available datasets: UCF-QNRF~\cite{idrees2018composition}, UCF CC~\cite{idrees2013multi}, Shanghaitech~\cite{zhang2016single} and NWPU~\cite{wang2020nwpu}. In \Sref{sec:Crowd Density Prediction}, we discuss the crowd counting results on all datasets. Note that we use the entire training set from each dataset for this experiment. In \Sref{sec:Uncertainty Guided Query Selection}, we show the effectiveness of the proposed sample selection strategies. Following previous works, we report Mean Absolute Error (MAE) and Root Mean Squared Error (RMSE) metrics:
\begin{eqnarray}
MAE = \frac{1}{n}\sum_{i=1}^{n} \lvert C_i - \hat{C}_i \rvert; 
RMSE = \sqrt[]{\frac{1}{n}\sum_{i=1}^{n} (C_i - \hat{C}_i)^2}, \nonumber 
\end{eqnarray}
where $C_i$ is the ground truth count, $\hat{C}_i$ is the predicted count, and the summation is computed over all test images. 

\subsection{Crowd Density Prediction\label{sec:Crowd Density Prediction}}

\subsubsection{Experiments on UCF-QNRF dataset.}

The UCF-QNRF dataset~\cite{idrees2018composition} consists of 1201 training and 334 test images of variable sizes, with 1.2 million dot annotations. For our experiments, we rescale those images for which the larger side is greater than 2048 pixels to 2048. For training, we take random crops of size $512{\times}512$ from each image.
Keeping the variance prediction branch fixed, we first train the other blocks of the proposed network for 20 epochs using the mean squared error loss. Next, we train only the uncertainty variance prediction head by minimizing the negative log likelihood objective for five epochs. Finally, we train the entire network by minimizing the negative log likelihood for 10 more epochs, and report the best results. We use a learning rate of $10^{-4}$, and a batch size of three for training. 

\myheading{Comparison with existing approaches.}
\Tref{tab:tableucf_nwpu} shows the performance of various approaches on the UCF-QNRF dataset. Bayesian Loss \cite{bayesianCounting} is a novel loss function for training crowd counting networks. It outperforms mean squared error, and it has the current state-of-the-art performance. This loss function is complimentary to what we propose here, and it can be used together with CTN. In fact, the method CTN$^{*}$ displayed in \Tref{tab:tableucf_nwpu} is the combination of CTN and Bayesian Loss.  CTN$^{*}$ improves the performance of Bayesian Loss \cite{bayesianCounting} and advances the state-of-the-art result on this dataset. 

\myheading{Ablation Study.} The proposed CTN consists of three main components: Local Feature Block, Non-Local Feature Block, and Predictive Uncertainty Estimation Branch.  To further understand the contribution of each component, we perform an ablation study, and the results are shown in  \Tref{tab:tableucf_ablation}. As can be seen, all constituent components of CTN are important for maintaining its good performance on the UCF-QNRF dataset. 

\setlength{\tabcolsep}{4pt}
\begin{table}[!tb]
\centering
\begin{tabular}{lcccc}
\toprule
         & \multicolumn{2}{c}{UCF-QNRF} & \multicolumn{2}{c}{NWPU} \\
         \cmidrule(lr){2-3} \cmidrule(lr){4-5} 
               & MAE          & RMSE         & MAE          & RMSE        \\
\midrule
Idrees \textit{et al.}~\cite{idrees2013multi} & 315 & 508 & - & - \\
MCNN~\cite{zhang2016single} & 277 & 426 & 219 & 701 \\
CMTL~\cite{sindagi2017cnn}& 252 & 514 & - & - \\
Switch CNN ~\cite{Sam-etal-CVPR17} & 228 & 445 & - & - \\
Composition Loss-CNN~\cite{idrees2018composition} & 132 & 191 & - & -  \\
CSR net~\cite{li2018csrnet} & - & -& 105 & 433\\
CAN ~\cite{liu2019context} & 107 & 183 & 94 & 490\\
SFCN ~\cite{wang2019learning} & 102  &   171 & - & -  \\
ANF~\cite{zhang2019attentional} & 110 & 174  & - & - \\
Bayesian Loss~\cite{bayesianCounting} & 89  & 155 & 93  & 470 \\
SCAR~\cite{gao2019scar} & - & - & 82 & \textbf{398} \\
\hline
CTN$^{*}$ (Proposed)  & \textbf{86} & \textbf{146} & \textbf{78} & 448  \\
\bottomrule
\end{tabular}
\vskip 0.15in
\caption{{\bf Performance of various methods on the UCF-QNRF test dataset and NWPU validation dataset}. Bayesian Loss is a recently proposed novel loss function for training a crowd counting network. CTN$^*$ is the method that combines CTN and Bayesian Loss, advancing the state-of-the-art performance in both MAE and RMSE metrics.
Following ~\cite{bayesianCounting}, we use the first four blocks from Vgg-19 as the backbone for local feature extraction.\label{tab:tableucf_nwpu}}
\end{table}


\setlength{\tabcolsep}{3pt}
\begin{table}[!tb]
\centering
\begin{tabular}{lgdgd}
\toprule
Components & \multicolumn{4}{c}{Combinations}   \\
\midrule
Local features  & \checkmark & \checkmark & \checkmark & \\
Non-Local features & \checkmark& \checkmark & & \checkmark \\
Predictive Uncertainty & \checkmark & & & \\
\hline 
MAE & 93 & 106 & 120 & 123 \\
RMSE & 166 &  185 & 218 & 206 \\
\bottomrule
\end{tabular}
\vskip 0.15in
\caption{{\bf Ablation study on UCF-QNRF}. CTN is the proposed counting network that consists of: Local Feature Block, Non-local Feature Block, and Uncertainty Prediction Branch. All three components are important for maintaining the good performance of CTN on this dataset. Note that ablation study is done using the Vgg-16 backbone. \label{tab:tableucf_ablation}}
\end{table}

\myheading{Experiments on NWPU-Crowd dataset.}
NWPU~\cite{wang2020nwpu} is the largest crowd counting dataset comprising of 5,109 crowd images taken from the web and video sequences, and over $2.1$ million annotated instances.
The ground truth for test images are not available, here we present the results on the validation set of NWPU in \Tref{tab:tableucf_nwpu}.
For this experiment, we use the pretrained CTN model from UCF-QNRF dataset and adapt the network on the NWPU dataset. Our proposed approach outperforms the previous methods.




\myheading{Experiments on UCF-CC dataset.}
The UCF-CC dataset~\cite{idrees2013multi} consists of 50 images collected form the web, and the count across the dataset varies between 94 and 4545. We use random crops of size $\frac{H}{3}{\times}\frac{W}{3}$ for training. Following  previous works, we perform 5-fold cross validation and report the average result in \Tref{tab:table_ucfcc_shanghaitech}. The proposed CTN with the Predictive Uncertainty (CTN) is comparable to other state-of-the-art approaches in both MAE and RMSE metrics. For all the approaches, the error on UCF CC dataset is higher compared to the other datasets since it has a small number of training samples. 

\setlength{\tabcolsep}{4pt}
\begin{table}[!tb]
\centering
\begin{tabular}{lcccccc}
\toprule
         & \multicolumn{2}{c}{UCF-CC} & \multicolumn{2}{c}{Shtech Part A} & \multicolumn{2}{c}{Shtech Part B} \\
         \cmidrule(lr){2-3} \cmidrule(lr){4-5} \cmidrule(lr){6-7}
               & MAE          & RMSE         & MAE          & RMSE     & MAE          & RMSE       \\
\midrule
Crowd CNN~\cite{zhang2015cross}   & -  & -    &        181.8      &  277.7           &   32.0           &  49.8           \\

MCNN~\cite{zhang2016single}           &    377.6    &  509.1 &     110.2     &    173.2         &  26.4            &  41.3           \\
Switching CNN~\cite{sam2017switching}  & -  & - & 90.4           &  135.0           &     21.6         &     33.4        \\
CP-CNN~\cite{sindagi2017generating} &295.8 & 320.9  & 73.6 & 106.4 &20.1 & 30.1  \\
IG-CNN~\cite{babu2018divide} & 291.4 & 349.4& 72.5 & 118.2 & 13.6 & 21.1 \\
ic-CNN~\cite{ranjan2018iterative} & 260.9 & 365.5  & 68.5 &116.2 & 10.7& 16.0\\
SANet ~\cite{cao2018scale} & 258.4 & 334.9 &  67.0 & 104.5 & 8.4 & 13.6 \\
CSR Net ~\cite{li2018csrnet} & 266.1 & 397.5 & 68.2 & 115.0 & 10.6 & 16.0 \\
PACNN ~\cite{shi2019revisiting}  & 241.7 & 320.7 & 62.4 & 102.0 & 7.6 & \textbf{11.8} \\
SFCN ~\cite{wang2019learning}  & 214.2  &   318.2 & 64.8  & 107.5  &7.6   &  13.0 \\
ANF~\cite{zhang2019attentional}  & 250.2 & 340.0 & 63.9 & \textbf{99.4} &  8.3 & 13.2 \\
Bayesian Loss~\cite{bayesianCounting}  & 229.3 & 308.2 & 62.8 & 101.8  & 7.7 & 12.7  \\
\hline
CTN (proposed) & \textbf{210.0} & \textbf{305.4} &  \textbf{61.5} & 103.4 & \textbf{7.5} & 11.9 \\
\bottomrule
\end{tabular}
\vskip 0.15in
\caption{{\bf Count errors of different methods on the UCF-CC dataset and Shanghaitech dataset.}  This dataset has two parts: Part A was collected from the web, and Part B was collected from the streets of Shanghai. The average ground truth crowd count for Part A is larger than that for Part B. We report both MAE and RMSE metrics. \label{tab:table_ucfcc_shanghaitech}
}
\end{table}
\myheading{Experiments on Shanghaitech dataset.}
The Shanghaitech dataset~\cite{zhang2016single} consists of two parts. Part A contains 482 images collected from the web, and Part B contains 716 images collected on the streets of Shanghai. The average ground truth crowd counts for Part A and Part B are 501 and 124, respectively. 
For training, we use random crops of size $\frac{H}{3}{\times}\frac{W}{3}$. 
Results are shown in \Tref{tab:table_ucfcc_shanghaitech}. The proposed approach outperforms all existing approaches in terms of MAE. Part A is more challenging with denser crowds than Part B. As a result, the average error of all the approaches on Part A is larger than those on Part B. 
Note that the CTN network in \Tref{tab:table_ucfcc_shanghaitech} is first trained on UCF-QNRF and later finetuned on Shanghitech dataset. This may not be a fair comparison for those approaches in \Tref{tab:table_ucfcc_shanghaitech} where the networks are trained from scratch on Shanghaitech dataset. Hence, for a more fair comparison, 
 we train the current state of the art model~\cite{bayesianCounting} on UCF-QNRF dataset first, and finetune it on Shanghaitech Part A dataset. We use the official implementation by the authors and use the hyper parameters reported by the authors~\cite{bayesianCounting}. This results in MAE/RMSE of 63.4/107.9 on the test set of ShanghaiTech Part A. Our CTN outperforms~\cite{bayesianCounting} in this experiment, with MAE/RMSE of 61.5/103.4 reported in \Tref{tab:table_ucfcc_shanghaitech}. This is a fair comparison since both methods are pretrained on UCF-QNRF, and later finetuned on Shanghaitech Part A.

\setlength{\tabcolsep}{1.5pt}
\begin{table}[!t]
\small 
\begin{center}	
\begin{tabular}{lcccccc}
\toprule
          &  \multicolumn{3}{c}{ Shtech Part A} &  \multicolumn{3}{c}{NWPU} \\
         \cmidrule(lr){2-4} \cmidrule(lr){5-7} 
    Selection approach & \#Train           & MAE          & RMSE  & \#Train        & MAE          & RMSE         \\
\midrule
None (Pretrained) & NA & 69.2 &  113.5 & NA & 118.4   & 632.3  \\
\hline
Random & 50 & 68.7 & 117.1 & 100 & 117.4 & 640.7 \\
Count & 50 & 67.3 & 107.4 & 100 & 107.9 & 458.8\\
Aleatoric Uncertainty & 50 & 62.9 & 108.1 & 100 & 104.9 & 522.1 \\
\textbf{\small{Density based Ensemble Disagr.}} & \textbf{50} & \textbf{61.4} & \textbf{105.5}  & 100 & 112.8 & 526.8 \\
{KL-Ensemble Disagreement} & 50 & 65.5 & 118.4 & 100 & 105.8 & 481.9 \\
\hline
Random & 100 & 65.5 & 125.5 & 500 & 96.7 & 539.4 \\
Count & 100 & 63.3 & 109.8 & 500 & 95.9 & 442.5 \\
\textbf{Aleatoric Uncertainty}  & 100 & 64.7& 107.8 & \textbf{500} & \textbf{81.5} & \textbf{313.7} \\
\small{{Density based Ensemble Disagreement}} & 100 & 62.2 & 109.6  & 500 & 90.0 & 438.6 \\
{KL-Ensemble Disagreement} & 100 & 62.1 & 103.3 & 500 & 95.6 & 511.3  \\
\midrule
Full dataset (previous best methods) & 300  & 62.8 & 99.4 & 3109 & 82 & 398 \\
Full dataset (CTN) & 300  & 61.5 & 103.4 & 3109 & 78.1 & 448.2 \\

\bottomrule
\end{tabular}
\end{center}
\vskip 0.15in
\caption{{\bf Comparing different strategies for selecting images for annotation.} We train the network on the UCF-QNRF dataset, and use it to select images from the NWPU and Shanghaitech train data for acquiring annotation. We compare the random selection baseline with the proposed uncertainty-guided selection strategy. For NWPU dataset, using just 500 training samples selected using our sampling strategy, we achieve state-of-the-art results compared to the previous state-of-art~\cite{gao2019scar}  trained on entire training set in terms of MSE. For Shanghaitech Part A, using just 50 labeled training samples, selected using our density based ensemble disagreement sampling strategy, we perform comparably to the state-of-the-art networks trained on the entire training set.  \label{query_shanghaitech_nwpu}}
\end{table}

\newcommand\imagescale{0.16}

\begin{figure*}[!h]
\begin{center}
\begin{tabular}{ccccc} 
{\bf Image} & {\bf Ground truth} & {\bf Mean} &  {\bf Variance} & {\bf Error}   \\ 
{  
	 \includegraphics[scale=\imagescale]{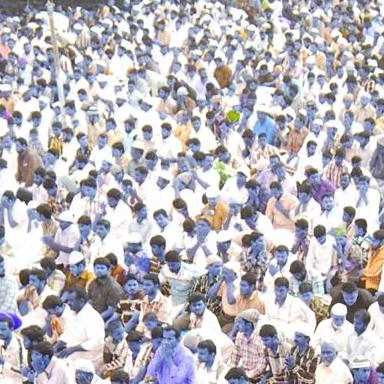}
     } &  {  
	\includegraphics[scale=\imagescale]{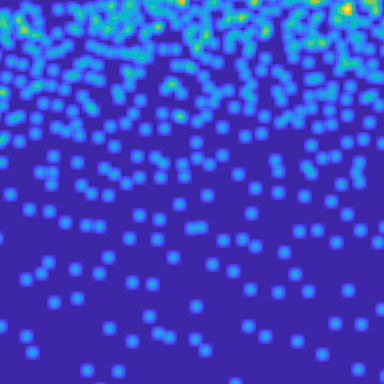}
     }
     &
     {\includegraphics[scale=\imagescale]{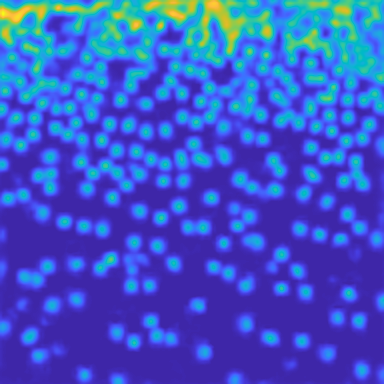}
     } 
          &
     {\includegraphics[scale=\imagescale]{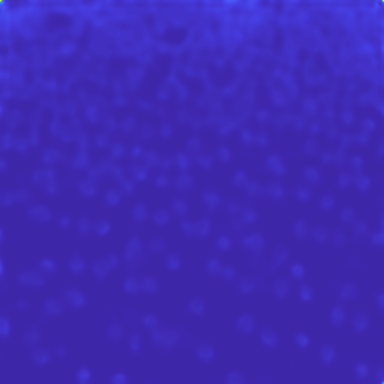}
     }
     &
        {\includegraphics[scale=\imagescale]{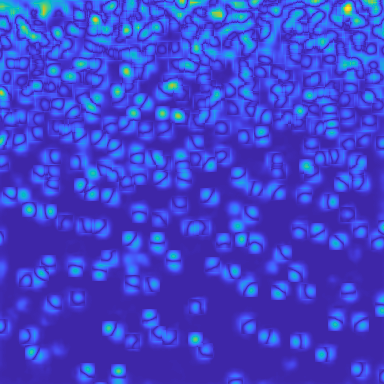}
     }
     \\
    &382 & 380 & 0.61 & 2 \\
   \\ 
{  
	\includegraphics[scale=\imagescale]{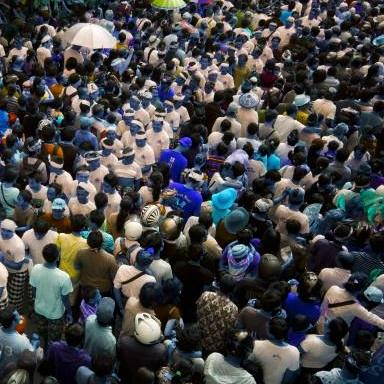}
     } &  {  
	\includegraphics[scale=\imagescale]{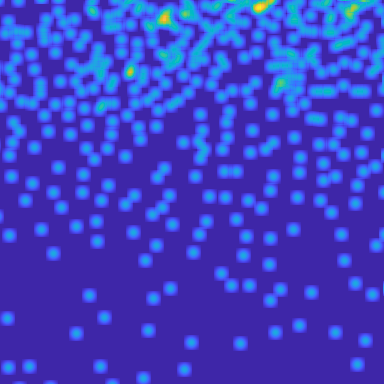}
     }
     &
     {\includegraphics[scale=\imagescale]{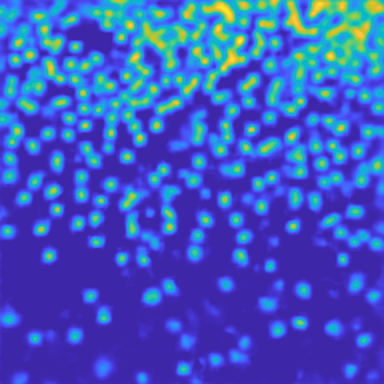}
     } 
          &
     {\includegraphics[scale=\imagescale]{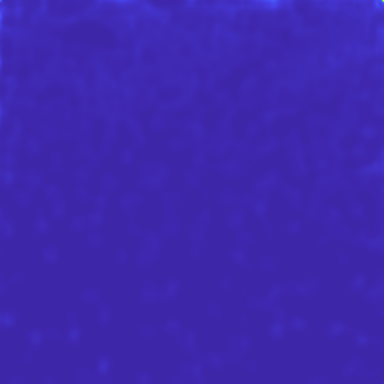}
     }
     &
        {\includegraphics[scale=\imagescale]{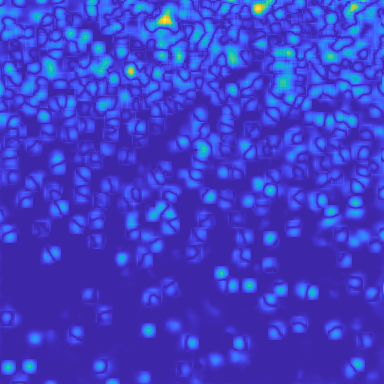}
     }
     \\
    &346 & 329 & 0.4 &17 \\
   \\ 

   {  
	\includegraphics[scale=\imagescale]{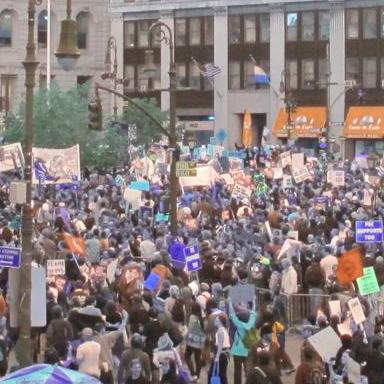}
     } &  {  
	\includegraphics[scale=\imagescale]{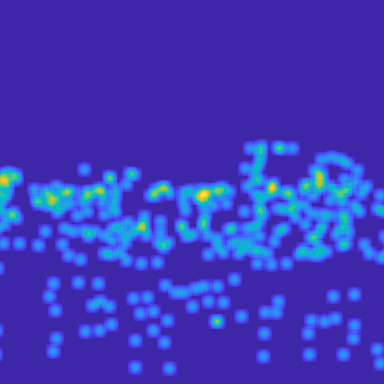}
     }
     &
     {\includegraphics[scale=\imagescale]{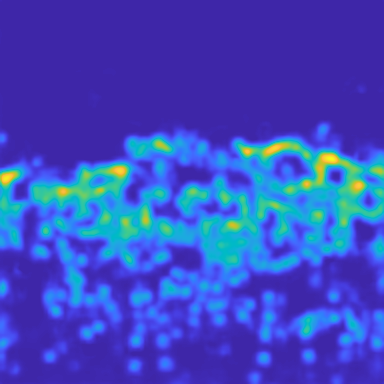}
     } 
          &
     {\includegraphics[scale=\imagescale]{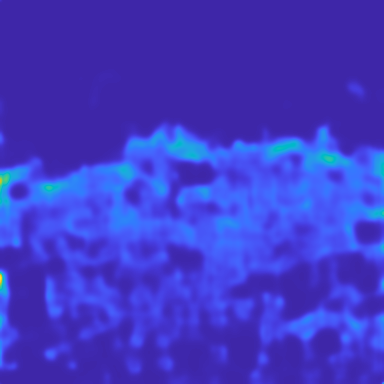}
     }
     &
        {\includegraphics[scale=\imagescale]{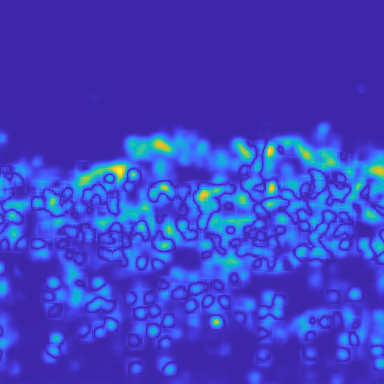}
     }
     \\
    &282 & 404 & 1.01 & 122\\
\end{tabular}
\caption{{\bf Qualitative Results.} This figure shows Image, Ground truth, Predicted Mean, Predicted Variance, and Error map. We specify the sum of the map below the corresponding map. The first two examples are success cases for density estimation, while the last is a failure cases. The variance maps correlate with the error maps. 
\label{fig:qualiContext}}
\end{center}
\end{figure*}

\subsection{Uncertainty Guided Image Annotation}\label{sec:Uncertainty Guided Query Selection}
In this section, we evaluate the effectiveness of the proposed selective annotation strategy. We train the network on the UCF-QNRF dataset and use it to select the informative samples from the Shanghaitech Part A dataset and NWPU dataset for acquiring annotation (results on Shanghaitech Part B are presented in the Supplementary). We use the labels of the selected samples, keep the variance prediction branch frozen, and finetune CTN using the selected subset. We compare our sampling approach with two baseline sampling approaches: 1) \textit{random sampling approach}: images are randomly sampled from the unlabeled pool in the target domain, and 2) \textit{Count based sampling}: we select those samples from the target domain for which the pretrained network predicts a high count. We report the results in \Tref{query_shanghaitech_nwpu}. Note that the entire training sets of Shanghaitech Part A and NWPU have  300 and 3109 images respectively. Our sampling approaches outperform the random baseline by a large margin. We also outperform the Count based sampling baseline. For NWPU dataset, using just 500 training samples, we achieve state-of-the-art results compared to the previous state-of-art~\cite{gao2019scar}  trained on the entire training set in terms of MSE. For Shanghaitech Part A, using just 50 labeled training samples, selected using our density based ensemble disagreement sampling strategy, we perform comparably to the state-of-the-art networks trained on the entire training set. Our results clearly show the usefulness of our informative sample selection strategy for transferring counting networks from one domain to another.

\subsection{Qualitative Results}
\Fref{fig:qualiContext} displays some qualitative585 results from UCF-QNRF dataset. Error is correlated with the variance which suggests the appropriateness of using the variance maps for estimating the uncertainty of the density prediction. 





\section{Conclusions}
To tackle large human annotation costs involved with annotating large scale crowd datasets, we have presented uncertainty based and ensemble disagreement based sampling strategies. These strategies were shown to be useful for the task of transferring a crowd network trained on one domain to a different target domain. Using just $17\%$ of the training samples obtained using our sampling strategy, we obtained 
state-of-the-art results on two challenging crowd counting datasets. We also showed that our proposed architecture, when trained on the full dataset, achieved state-of-the-art results on all the datasets in terms of mean absolute error.

\myheading{Acknowledgements}: This project is partially supported by MedPod and the SUNY2020 Infrastructure Transportation Security Center.

\newpage
\section{Supplementary Materials} 
\subsection{Overview}
 Earlier works~\cite{kendall2017uncertainties} have shown the usefulness of both Aleatoric and Epistemic Uncertainty estimates for various Computer Vision tasks. In the main paper, we presented our CTN architecture for estimating aleatoric uncertainty pertaining to crowd density prediction. In \Sref{sec:Bayesian Uncertainty Estimation}, we present Monte Carlo CTN (MC-CTN), our architecture for estimating  epistemic uncertainty. Subsequently in \Sref{sec:Evaluating Uncertainty Estimate}, we compare epistemic and aleatoric uncertainty estimates, and show that aleatoric uncertainty is more correlated with the prediction error. In the main paper, we focused on using aleatoric uncertainty estimates for sample selection since our preliminary experiments showed Aleatoric uncertainty to be more effective at sample selection compared with epistemic uncertainty. 
 
  In the main paper, we pointed out that our sampling strategy can be used for picking out images from a large pool of unlabeled images, and it can also be used to pick out informative crops from an image. Depending on the annotation budget, it might be useful to get partial annotations for an image by picking out informative crops from an image and getting human annotations for the informative crops rather than annotating the entire image. The experiments on image level sample selection are discussed in the main paper. In \Sref{sec:Crop-level Sample Selection}, we present experiments pertaining to selecting most informative crops from an image for human annotation.
  
  Finally, for image level sample selection, in addition to experiments performed on Shanghaitech Part A~\cite{zhang2016single} and NWPU~\cite{wang2020nwpu} datasets as shown in the main paper, we present more experiments on Shanghaitech Part B in \Sref{sec:PartB}.

\begin{figure}[!h]
\centering
\includegraphics[width=0.48\linewidth,height=0.4\linewidth]{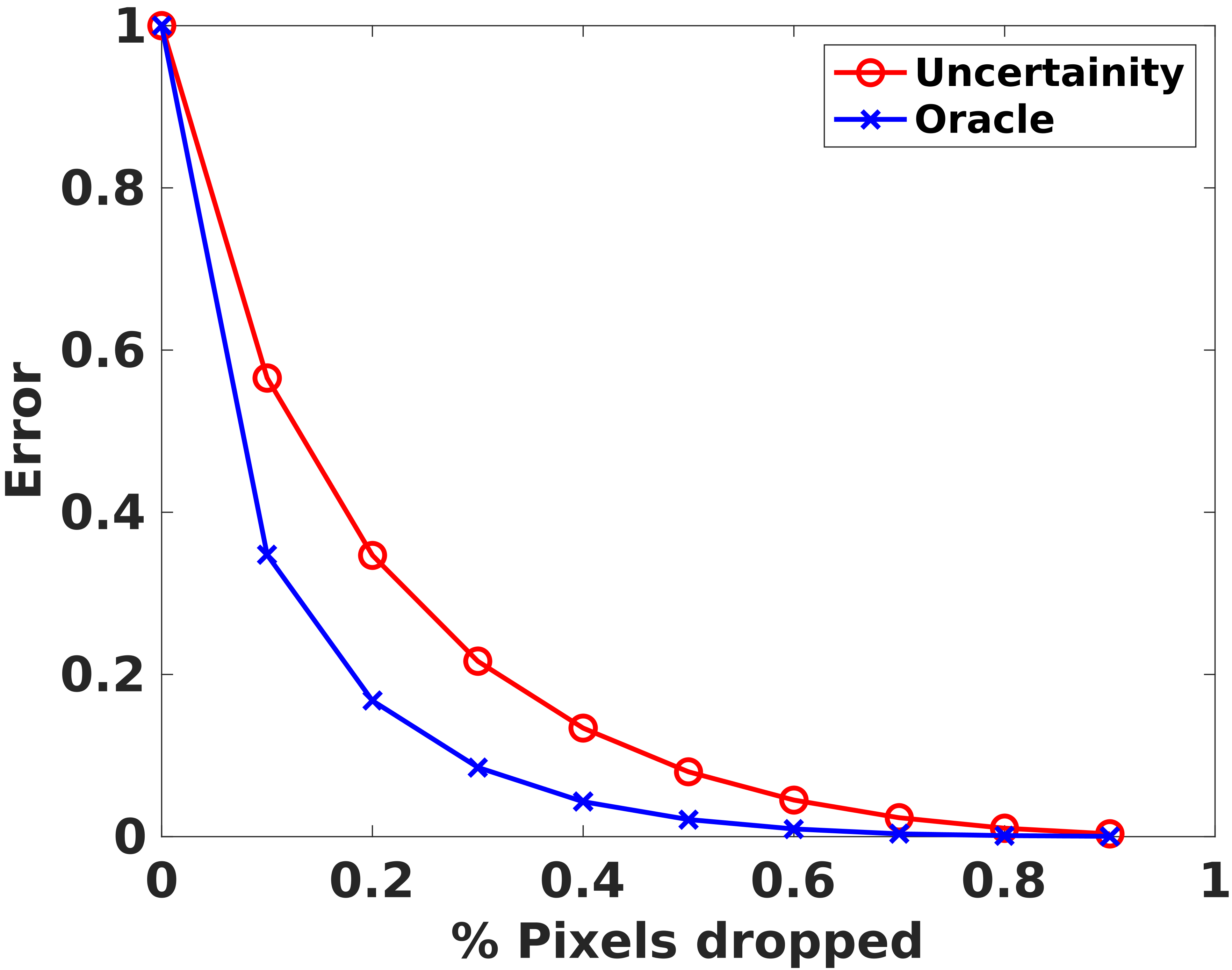}
\includegraphics[width=0.48\linewidth,height=0.4\linewidth]{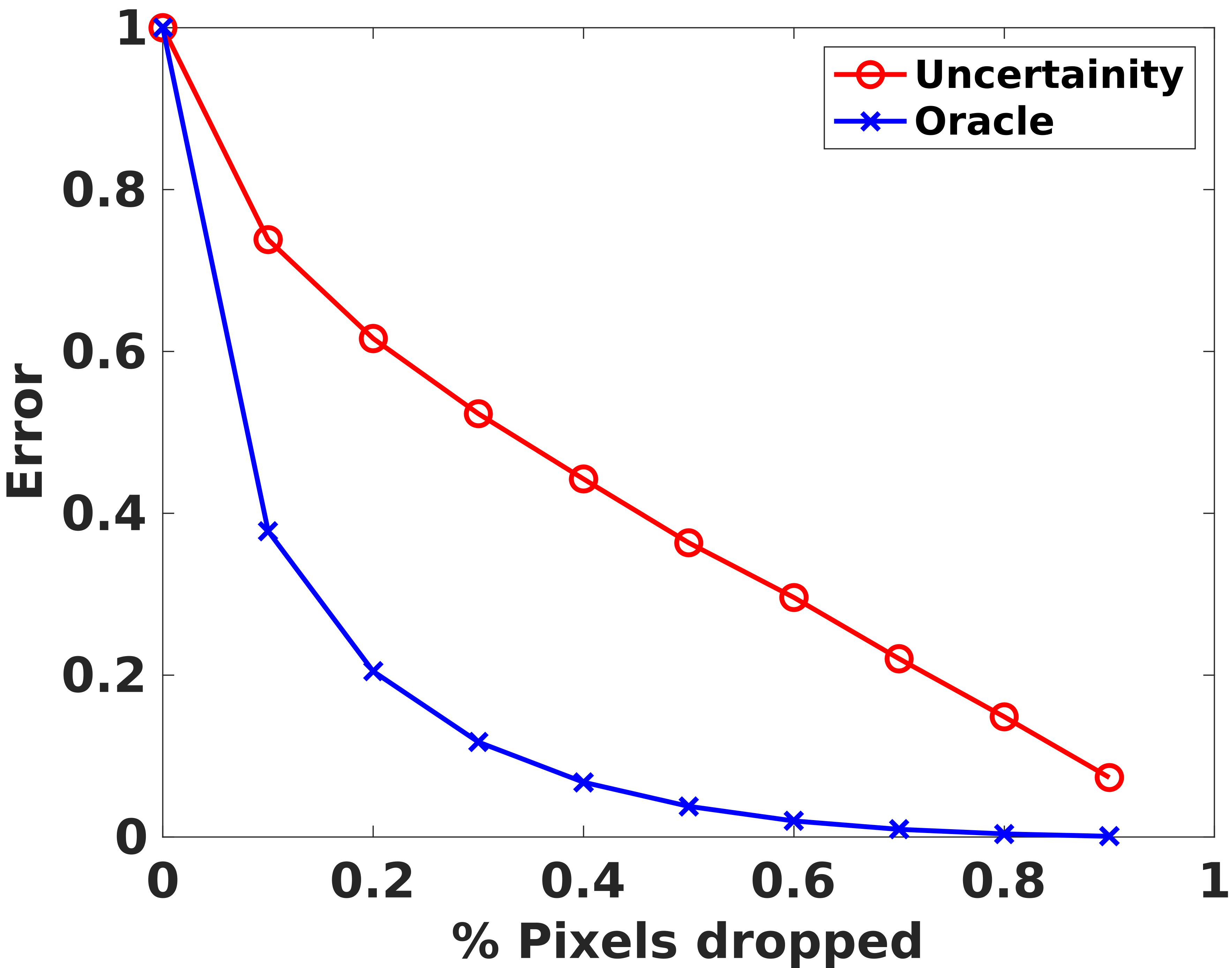}
	\caption{The sparsification plots for aleatoric (left) and epistemic (right) uncertainty estimation on the test set of Shanghaitech Part A. The closer the uncertainty curve to the oracle, the higher the correlation between the uncertainty and prediction error. The area  between the uncertainty curve and the oracle curve for the two cases is 0.07 and 0.25. The aleatoric uncertainty is more correlated with the prediction error. \label{fig:SparsificationPlot}}	
\end{figure}
\subsection{Epistemic Uncertainty Estimation}\label{sec:Bayesian Uncertainty Estimation}

The CTN architecture described in the main paper captures the aleatoric uncertainty, i.e., the uncertainty inherent in the input data. In this section, we present a variant of CTN which can capture the model uncertainty, which is also known as Epistemic Uncertainty and arises due to the use of finite training data. 
Epistemic uncertainty, can be captured by Bayesian Neural Networks~\cite{neal2012bayesian}. Such networks assume a prior distribution $\mathbb{P}(\theta)$ over the parameters of the network, and find the posterior probability $\mathbb{P}(\theta|\{X,Y\})$ conditioned on the training data $\{X,Y\}$. However, it is computationally expensive to perform inference with Bayesian Neural Networks with large number of parameters. To address this issue, \cite{gal2015bayesian} present a variational approximation which  corresponds to adding a dropout~\cite{srivastava2014dropout} layer before every weight layer. The uncertainty in such a network can be obtained by doing multiple forward passes of the input image through the network, and computing the sample mean as the prediction output. The sample variance is the epistemic uncertainty of the prediction. This approach requires running the
network multiple times with different dropout instantiations, so we will refer to it as Monte Carlo Uncertainty. 

We will refer to a counting network where Predictive Uncertainty is replaced by Monte Carlo Uncertainty as Monte Carlo CTN (MC-CTN). MC-CTN is similar to CTN presented in the main paper with two major differences: 1) we remove the Predictive Uncertainty Estimation branch, and 2) we add a dropout layer before every convolution layer except for the ones in density prediction branch.

\myheading{Performance of MC-CTN on Crowd Counting}
MC-CTN led to MAE/RMSE of 102/180 on UCF-QNRF dataset. CTN, as reported in the main paper, outperforms MC-CTN and results in MAE/RMSE of 86/146.

\setlength{\tabcolsep}{1.5pt}
\begin{table}[t]
\small 
\begin{center}	
\begin{tabular}{lccc}
\toprule
          &  \multicolumn{3}{c}{ Shtech Part A}  \\
         \cmidrule(lr){2-4} 
    Selection approach & \#Crops           & MAE          & RMSE          \\
\midrule
None (Pretrained) & NA & 69.2 & 113.5  \\
\hline
Random & 1 of 16  & 68.6 &   113.5  \\
Count & 1 of 16   & 68.2 & 113.9  \\
\textbf{\small{Aleatoric Uncertainty}} & 1 of 16   & 65.7 &   103.5  \\
Density based Ensemble Disagre.   & 1 of 16  & 66.4 &  112.4  \\
{KL-Ensemble Disagreement} & 1 of 16   &66.4  &   109.7  \\
\midrule
Full dataset (previous best method) & full image   & 62.8 &    99.4\\
Full dataset (CTN) & full image   & 61.5 & 103.4   \\

\bottomrule
\end{tabular}
\end{center}
\vskip 0.15in
\caption{{\bf Comparing different strategies for selecting informative crops from an image for annotation.} We train the network on the UCF-QNRF dataset, and use it to select informative crops Shanghaitech Part A train data for acquiring annotation. The selected crops are used to adapt the network to the target domain. We compare the random sample selection baselines with the proposed uncertainty-guided selection strategies. For the experiment, each image is divided into 16 crops and a single crop from each image is chosen for annotation. We compute
the informativeness score for all 16 crops, and pick out the crop with the highest score. For the random baseline, we pick out a crop at random from each image. For the count baseline, we pick out the crop with the highest count. Our proposed sample selection strategies outperforms the random selection and count based selection baselines. \label{cropSelection}}
\end{table}

\subsection{Comparing Aleatoric and Epistemic Uncertainties}\label{sec:Evaluating Uncertainty Estimate}
Sparsification plots provide a way to ascertain whether the uncertainty estimate is correlated with the prediction error~\cite{ilg2018uncertainty}.  Such plots are obtained by removing pixels with the highest uncertainty and measuring the error for the remaining pixels. If the predicted uncertainty is correlated with the error, the overall error should monotonically decrease as we remove more uncertainty pixels. We show the sparsification plots for the Aleatoric and Epistemic uncertainty estimates in \Fref{fig:SparsificationPlot}. The optimal uncertainty estimate, would be perfectly correlated with the error, and the characteristic curve for the best possible uncertainty estimate can be obtained by dropping pixels with the highest errors first.
We refer to this curve as {\it Oracle} in \Fref{fig:SparsificationPlot}, and the closer the uncertainty sparsification curve to the Oracle curve, the better.
The area between the sparsification curve and the oracle curve  for the Aleatoric and Epistemic uncertainty estimations are 0.07 and 0.25, respectively. This shows that the prediction error has higher correlation with the aleatoric uncertainty estimation than the epistemic uncertainty estimation. Similar evaluation strategies have been used for evaluating the uncertainty estimates pertaining to optical flow estimation~\cite{ilg2018uncertainty}.
\subsection{Crop-level Sample Selection}\label{sec:Crop-level Sample Selection}

 In the main paper, we pointed out that our sampling strategy can be used for picking out images from a large pool of unlabeled images, and it can also be used to pick out informative crops from an image. Depending on the annotation budget, it might be useful to get partial annotations for an image by picking out informative crops from an image and getting human annotations for the informative crops rather than annotating the entire image. The experiments on image level sample selection are discussed in the main paper. In \Tref{cropSelection}, we present experiments pertaining to selecting informative crops from an image for human annotation.
 For the experiment, each image is divided into 16 non-overlapping crops and a single crop from each image is chosen for annotation. We compute the informativeness score for all 16 crops, and pick out the crop with the highest score. The informativeness score of crops are computed using the three sample selection strategies proposed in the main paper. We compare our proposed approach with two baselines: random crop selection and count based crop selection. For the random baseline, we pick out a crop at random from each image. For the count baseline, we pick out the crop with the highest count. Our proposed sample selection strategies outperforms the random selection and count selection baselines.

 \subsection{Image-level Sample Selection on Shanghaitech Part B}\label{sec:PartB}

 In the main paper, we presented image level sample selection experiments on Shanghaitech Part A~\cite{zhang2016single} and NWPU~\cite{wang2020nwpu} datasets. In \Tref{PartB}, we present image level sample selection experiments on Shanghaitech Part B. 
 
\setlength{\tabcolsep}{1.5pt}
\begin{table}[!b]
\small 
\begin{center}	
\begin{tabular}{lccc}
\toprule
          &  \multicolumn{3}{c}{ Shtech Part B}  \\
         \cmidrule(lr){2-4} 
    Selection approach & \#Train           & MAE          & RMSE          \\
\midrule
None (Pretrained) & NA & 13.2 & 21.7  \\
\hline
Random & 50  & 9.4 & 16.6    \\
Count & 50  & 8.9 & 15.1  \\
Aleatoric Uncertainty & 50  & 8.7 & 14.3    \\
Density based Ensemble Disagr.   & 50 & 8.5 & 13.6   \\
{KL-Ensemble Disagreement} & 50  & 8.8 & 15.4    \\
\hline
Random & 100  & 9.1 & 15.1   \\
Count & 100  & 8.0 &    13.5 \\
Aleatoric Uncertainty & 100  & 8.8 & 15.0    \\
\small{{Density based Ensemble Disagreement}} & 100  & 8.5 &  14.5  \\
{KL-Ensemble Disagreement} & 100  & 8.5 & 13.7      \\
\midrule
Full dataset (previous best method)~\cite{shi2019revisiting} & 400   & 7.6 &    11.8\\
Full dataset (CTN) & 400   & 7.5 & 11.9   \\

\bottomrule
\end{tabular}
\end{center}
\vskip 0.15in
\caption{{\bf Comparing different strategies for selecting images for annotation on Shanghaitech Part B dataset} We train the
network on the UCF-QNRF dataset, and use it to select images from the Shanghaitech Part B train data for acquiring annotation. We compare the random selection baseline with the proposed uncertainty-guided selection strategy.  Our sample selection strategies outperform the two baselines when 50 images are sampled, and outperforms the random baseline when sampling 100 images.  \label{PartB}}
\end{table}

\setlength{\bibsep}{0pt} 
\bibliographystyle{plainnat}
\bibliography{longstrings,egbib,pubs}

\end{document}